\documentclass[12pt]{article}

\usepackage[utf8]{inputenc}
\usepackage{graphicx}
\usepackage{amsmath}
\usepackage{amsfonts}
\usepackage{amssymb}
\usepackage{geometry}
\usepackage{float}
\usepackage{caption}
\usepackage{subcaption}
\usepackage{booktabs}
\usepackage{array}
\usepackage{multirow}
\usepackage{longtable}
\usepackage{setspace}
\usepackage{url}
\usepackage{cite}
\usepackage{hyperref}
\usepackage{enumitem}
\usepackage{parskip}

\graphicspath{{images/media/}{figures/}}
\DeclareGraphicsExtensions{.pdf,.png,.jpg,.jpeg}

\hypersetup{
    colorlinks=true,
    linkcolor=blue,
    filecolor=magenta,      
    urlcolor=cyan,
    citecolor=blue,
    pdfborder={0 0 0}
}

\begin{document}

\title{\Large \textbf{A Novel Evaluation Benchmark for Medical LLMs: Illuminating Safety and Effectiveness in Clinical Domains}}

\author{%
\begin{minipage}{\textwidth}\centering\large
\mbox{Shirui~Wang\textsuperscript{1\#}}, \mbox{Zhihui~Tang\textsuperscript{2\#}}, \mbox{Huaxia~Yang\textsuperscript{3\#}}, \mbox{Qiuhong~Gong\textsuperscript{4\#}}, \mbox{Tiantian~Gu\textsuperscript{1\#}}, 
\mbox{Hongyang~Ma\textsuperscript{2\#}}, \mbox{Yongxin~Wang\textsuperscript{1}}, \mbox{Wubin~Sun\textsuperscript{1}}, \mbox{Zeliang~Lian\textsuperscript{1}}, \mbox{Kehang~Mao\textsuperscript{1}}, 
\mbox{Yinan~Jiang\textsuperscript{5}}, \mbox{Zhicheng~Huang\textsuperscript{6}}, \mbox{Lingyun~Ma\textsuperscript{7}}, \mbox{Wenjie~Shen\textsuperscript{8}}, \mbox{Yajie~Ji\textsuperscript{9}}, 
\mbox{Yunhui~Tan\textsuperscript{10}}, \mbox{Chunbo~Wang\textsuperscript{11}}, \mbox{Yunlu~Gao\textsuperscript{12}}, \mbox{Qianling~Ye\textsuperscript{13}}, \mbox{Rui~Lin\textsuperscript{14}}, 
\mbox{Mingyu~Chen\textsuperscript{15}}, \mbox{Lijuan~Niu\textsuperscript{16}}, \mbox{Zhihao~Wang\textsuperscript{17}}, \mbox{Peng~Yu\textsuperscript{18}}, \mbox{Mengran~Lang\textsuperscript{17}}, 
\mbox{Yue~Liu\textsuperscript{17}}, \mbox{Huimin~Zhang\textsuperscript{19}}, \mbox{Haitao~Shen\textsuperscript{20}}, \mbox{Long~Chen\textsuperscript{21}}, \mbox{Qiguang~Zhao\textsuperscript{22}}, 
\mbox{Si-Xuan~Liu\textsuperscript{9}}, \mbox{Lina~Zhou\textsuperscript{23}}, \mbox{Hua~Gao\textsuperscript{1}}, \mbox{Dongqiang~Ye\textsuperscript{1}}, \mbox{Lingmin~Meng\textsuperscript{1}}, 
\mbox{Youtao~Yu\textsuperscript{24*}}, \mbox{Naixin~Liang\textsuperscript{6*}}, \mbox{Jianxiong~Wu\textsuperscript{17*}}
\end{minipage}}

\date{}

\maketitle

\begin{flushleft}
\scriptsize
\textsuperscript{1}Medlinker Intelligent and Digital Technology Co., Ltd, Beijing, China\\
\textsuperscript{2}Peking University School of Stomatology, Beijing, China\\
\textsuperscript{3}Dept. of Rheumatology and Clinical Immunology, Peking Union Medical College Hospital, Chinese Academy of Medical Sciences and Peking Union Medical College, Beijing, China\\
\textsuperscript{4}Center of Endocrinology, National Center of Cardiology \& Fuwai Hospital, Chinese Academy of Medical Sciences and Peking Union Medical College, Beijing, China\\
\textsuperscript{5}Dept. of Psychological Medicine, Peking Union Medical College Hospital, Chinese Academy of Medical Sciences and Peking Union Medical College, Beijing, China\\
\textsuperscript{6}Dept. of Thoracic Surgery, Peking Union Medical College Hospital, Chinese Academy of Medical Sciences and Peking Union Medical College, Beijing, China\\
\textsuperscript{7}Dept. of Respiratory and Critical Care Medicine, the 8th Medical Center of PLA General Hospital, Beijing, China\\
\textsuperscript{8}Dept. of Obstetrics \& Gynecology, the Fourth Medical Center of PLA General Hospital, Beijing, China\\
\textsuperscript{9}Shuguang Hospital Affiliated to Shanghai University of Traditional Chinese Medicine, Shanghai, China\\
\textsuperscript{10}Dept. of Urology, The Second Affiliated Hospital of Harbin Medical University, Heilongjiang, China\\
\textsuperscript{11}Dept. of Radiation Oncology, Harbin Medical University Cancer Hospital, Harbin, Heilongjiang, China\\
\textsuperscript{12}Dept. of Dermatology, Shanghai Skin Disease Hospital, Tongji University School of Medicine, Shanghai, China\\
\textsuperscript{13}Dept. of Oncology, East Hospital Affiliated to Tongji University, Shanghai, China\\
\textsuperscript{14}General Surgery Dept., Tongji Hospital, School of Medicine, Tongji University, Shanghai, China\\
\textsuperscript{15}Dept. of Neurosurgery, Huashan Hospital, Shanghai Medical College, Fudan University, Shanghai, China\\
\textsuperscript{16}Dept. of Ultrasound, National Cancer Center/Cancer Hospital, Chinese Academy of Medical Sciences and Peking Union Medical College, Beijing, China\\
\textsuperscript{17}Dept. of Hepatobiliary Surgery, National Cancer Center/Cancer Hospital, Chinese Academy of Medical Sciences and Peking Union Medical College, Beijing, China\\
\textsuperscript{18}Dept. of General Surgery, The Fourth Affiliated Hospital of Xinjiang Medical University, Urumqi, China\\
\textsuperscript{19}Dept. of Otolaryngology-Head and Neck Surgery, Shanxi Bethune Hospital, Shanxi Academy of Medical Sciences, Taiyuan, Shanxi, China\\
\textsuperscript{20}Dept. of Clinical Laboratory, Seventh People's Hospital of Shanghai University of Traditional Chinese Medicine\\
\textsuperscript{21}Dept. of Orthopedics, Guangzhou Red Cross Hospital of Jinan University, Guangzhou, China\\
\textsuperscript{22}Dept. of Imageology, Anzhen Hospital, Capital Medical University, Beijing, China\\
\textsuperscript{23}Beijing EuroEyes, Beijing, China\\
\textsuperscript{24}Dept. of Interventional Radiology, the Fourth Medical Center of Chinese PLA General Hospital, Beijing, China\\[1em]

\small
\textbf{*Corresponding authors:}\\
Youtao Yu (yuyoutao@126.com),\\
Naixin Liang (pumchnelson@163.com),\\
Jianxiong Wu (Dr\_wujx@163.com)\\[0.5em]

\textsuperscript{\#}These authors contributed equally to this work
\end{flushleft}

\begin{abstract}
Large language models (LLMs) hold promise in clinical decision support but face major challenges in safety evaluation and effectiveness validation. We developed the Clinical Safety-Effectiveness Dual-Track Benchmark (CSEDB), a multidimensional framework built on clinical expert consensus, encompassing 30 criteria covering critical areas like critical illness recognition, guideline adherence, and medication safety, with weighted consequence measures. Thirty-two specialist physicians developed and reviewed 2,069 open-ended Q\&A items aligned with these criteria, spanning 26 clinical departments to simulate real-world scenarios. Benchmark testing of six LLMs revealed moderate overall performance (average total score 57.2\%, safety 54.7\%, effectiveness 62.3\%), with a significant 13.3\% performance drop in high-risk scenarios (p $<$ 0.0001). Domain-specific medical LLMs showed consistent performance advantages over general-purpose models, with relatively higher top scores in safety (0.912) and effectiveness (0.861). The findings of this study not only provide a standardized metric for evaluating the clinical application of medical LLMs, facilitating comparative analyses, risk exposure identification, and improvement directions across different scenarios, but also hold the potential to promote safer and more effective deployment of large language models in healthcare environments.
\end{abstract}

\section{Introduction}

The application of large language models (LLMs) in the medical domain is
advancing rapidly, generating broad research interest in the field of
AI-driven digital medicine\cite{ref1,ref2,ref3}. These models have
demonstrated significant potential to improve healthcare
outcomes\cite{ref4}. Representative LLMs such as ChatGPT and
DeepSeek-R1, with their powerful natural language processing and
reasoning capabilities, are expected to enhance the quality and
efficiency of healthcare services\cite{ref5}. For instance, LLMs
can help alleviate the strain on healthcare resources by providing
preliminary analyses of patient symptoms and answering common questions;
the patient-friendly medical information they generate can also improve
patient understanding of their conditions and treatment
regimens\cite{ref6,ref7}. Additionally, by serving as auxiliary
tools to address patient queries, they can facilitate better
physician--patient communication\cite{ref8}. However,
significant vulnerabilities in the safety and effectiveness of these
AI-driven platforms remain. In particular, LLMs can produce erroneous or
inaccurate information in medical outputs, posing potential risks to
patient health\cite{ref9,ref10}. Therefore, establishing robust
evaluation frameworks to validate their clinical applicability,
particularly with respect to safety and effectiveness, has become a
central challenge in digital medicine.

Current assessments of the clinical capabilities of LLMs primarily rely
on standardized medical examinations such as USMLE-style tests and
specialized QA datasets\cite{ref11,ref12} Yet strong performance on
such examinations does not necessarily equate to reliable deployment in
real-world clinical practice, where more comprehensive ``field testing''
is required\cite{ref3}. In the domain of safety evaluations,
several representative studies have emerged:
SafeBench\cite{ref13} focuses on multimodal LLMs, simulating
diverse scenarios to detect vulnerabilities arising from cross-modal
inputs; Agent-SafetyBench\cite{ref14} targets LLM-based agents
by identifying risks in their decision-making logic and behavioral
outputs; and aiXamine\cite{ref15} serves as a black-box
evaluation platform integrating over 40 tests, encompassing general
safety as well as healthcare-specific safety dimensions. However, many
of these approaches remain grounded in physician licensing examination
questions. Although they capture factual knowledge and reasoning
capabilities, they fail to comprehensively evaluate clinical practice
readiness\cite{ref3}. Fragmented evaluation dimensions that
overly emphasize performance on specific tasks, such as diagnostic
accuracy lack systemic analysis of the safety--effectiveness interplay,
potentially obscuring systemic risks in complex clinical
contexts\cite{ref16}. The absence of evidence-based risk
stratification standards can lead to fatal errors and hinder targeted
model optimization\cite{ref17}. In addition, insufficient
contextualization for real-world clinical settings fails to meet the
needs of special populations, such as pediatric dose calculation and the
time-sensitive demands of critical care, creating a translational gap
between technical validation and clinical application. Finally,
evaluation methods relying heavily on human assessors suffer from
subjectivity and low reproducibility, severely limiting
scalability\cite{ref18}. These compounded limitations underscore
the urgent need for a multidimensional evaluation framework that can
establish actionable mappings between technical metrics and dynamic
clinical realities.

Within current evaluation methodologies for medical LLMs,
question-and-answer (QA) formats remain the most common and can be
divided into closed-ended and open-ended tasks. Closed-ended tasks
evaluate specific model capabilities within a predefined answer space,
most commonly through multiple-choice questions (MCQs), as exemplified
by datasets such as MedQA, PubMedQA, and MedMCQA\cite{ref19}.
These tasks are readily standardized, as performance can be quantified
by answer accuracy without requiring continuous expert oversight. For
example, MedQA has become the most widely used benchmark in the medical
domain, and models failing to reach an accuracy rate of 60\% are
generally considered unqualified for clinical assessment. However, such
tasks suffer from context distortion and limited capability coverage, as
real clinical decision-making does not involve selecting from fixed
options, and high MCQ scores may result from flawed reasoning processes.
Open-ended tasks, by contrast, focus on the multidimensional quality of
model outputs, such as generating free-text diagnostic plans or
interpreting complex medical records. The MultiMedQA dataset, for
instance, was used in Med-PaLM\cite{ref19} evaluations to
represent these scenarios, offering greater alignment with real-world
clinical needs. Nevertheless, traditional natural language generation
(NLG) metrics correlate poorly with expert judgment, and the high cost
and low scalability of human assessments remain significant barriers.
Recent studies such as CRAFT-MD\cite{ref20},
AMIE\cite{ref21}, and AgentClinic\cite{ref22} have
explored new directions for open-ended evaluation by simulating
interactions between AI agents and LLMs. In addition, some recent
research\cite{ref23,ref24,ref25} has proposed leveraging patient
simulators to achieve automated evaluations based on predefined clinical
skills.

To address the compounded limitations of existing evaluation frameworks,
this study proposes a multidimensional evaluation framework driven by
clinical risk. We adopt a rubric-based evaluation approach that
integrates expert-defined assessment criteria with automated batch
testing to balance evaluation accuracy and efficiency, building on the
successful implementation of OpenAI's healthBench and related work in
medical settings\cite{ref17,ref25}. Specifically, we established an
open-ended QA framework that encompasses 26 clinical departments and 30
assessment criteria, including 17 safety-focused and 13
effectiveness-focused indicators. For the first time, this framework
enables standardized, two-dimensional benchmark of LLM performance in
terms of safety and effectiveness. The resulting benchmark provides a
scientific basis for model optimization and regulatory approval and
paves the way for the safe and effective translation of LLMs from
controlled laboratory environments to real-world clinical practice.

\section{Results}

\subsection{Research Design}

To evaluate the clinical utility of LLMs in consultation settings, we
designed the Clinical Safety-Effectiveness Dual-Track Benchmark (CSEDB).
This framework focuses on two core dimensions: safety (encompassing
critical illness detection and medication safety) and effectiveness
(encompassing guideline adherence and optimization of diagnostic and
therapeutic pathways). It aims to dissect the key capability elements
that influence patient outcomes when LLMs are used to assist clinical
decision-making. Priority was given to selecting assessment criteria
that are both technically compatible with the interactive reasoning
patterns of LLMs and directly linked to real-world clinical risks, such
as stratification of critical illness risk during dialogue or alerts for
potential drug--drug interactions.

Based on consensus among clinical experts regarding the relationship
between each indicator and its associated clinical risks and benefits,
we established 30 assessment criteria that cover critical illness
recognition, guideline adherence, medication safety, and optimization of
treatment strategies (Supplementary Table S1). Using these indicators as
the foundation and reflecting the complexity of real-world clinical
cases, we synthesized 2,069 clinical scenario questions encompassing 26
specialties and diverse patient populations, including elderly patients
with polypharmacy and individuals with immunodeficiency. Each question
was reviewed and validated by a panel of 32 specialist physicians, who
also developed standardized evaluation criteria for the responses to
each scenario (Supplementary Table S2).

The two-dimensional quantitative evaluation system adopts a hybrid
methodology that integrates binary classification and graded scoring.
Within the 17 safety-related dimensions, eight absolute contraindication
scenarios, such as the use of codeine in pediatric patients or
administration of aminoglycosides in patients with an estimated
glomerular filtration rate (eGFR$<$30), were assessed using binary
classification (safe versus unsafe). The remaining nine safety-related
scenarios, which require comprehensive clinical judgment, such as
antihypertensive dose adjustments in patients with chronic kidney
disease or stratification of drug--drug interaction risks in
polypharmacy, were assessed using graded scoring based on the
completeness of risk control. For the 13 effectiveness-related
dimensions, five scenarios involving explicit guideline-prohibited
practices, such as the overuse of magnetic resonance imaging in
nonspecific low back pain, were evaluated using binary classification
(appropriate versus inappropriate), while the remaining eight scenarios
requiring multidimensional evaluation, such as the strength of evidence
supporting targeted therapy regimens in oncology or the degree of
empathy expressed during clinician--patient communication, were
evaluated using graded scoring based on diagnostic and therapeutic value
and patient benefit.

The final scores were derived by weighting and normalizing all safety
and effectiveness indicators, with higher scores indicating stronger
alignment with best clinical practices. This evaluation methodology
combines automated assessment with manual concordance validation to
ensure robustness (Figure 1). As this study did not involve the
collection of patient data or any patient interventions, all procedures
complied with the Declaration of Helsinki.

\begin{figure}[htbp]
  \centering
  \includegraphics[width=5.76806in,height=2.4354in]{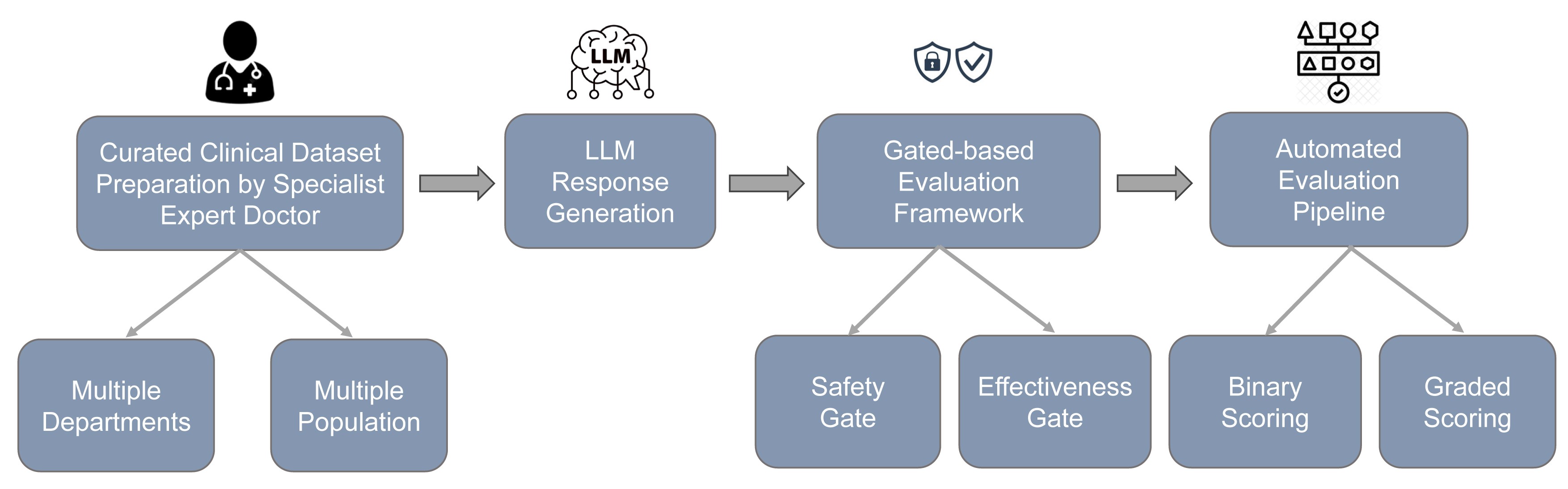}
  \caption{Overall research design workflow.}
  \label{fig:image1}
\end{figure}

\subsection{Core Performance Comparison: Overall Model Scores on Safety and
Effectiveness}

To investigate the performance of various LLMs on CSEDB framework, we
employed Deepseek-R1-0528, OpenAI-o3 (20250416), Gemini-2.5-Pro
(20250506), Qwen3-235B-A22B, Claude-3.7-Sonnet (20250219) and MedGPT
(MG-0623, Medlinker) as the test models. All evaluations were conducted
within a comparable time window, specifically between May 2025 and June
2025. Although our dataset primarily targets Chinese medical questions,
all the included models were trained predominantly on English data.
During the experiments, models were sampled at a temperature of 1.0,
while all other parameters were kept at their default configurations.

From the overall evaluation scores, the average performance across all
LLMs was 57.2\% ± 24.5\%, suggesting that their usability in clinical
settings remains at a moderate level. Performance in safety (average
54.7\% ± 26.1\%) was lower than that in effectiveness (average 62.3\% ±
22.3\%). The domain-specific medical model MedGPT outperformed the
general-purpose LLMs by a substantial margin, scoring 15.3\% higher than
the second-best model overall and 19.8\% higher in the safety dimension.
These findings indicate that MedGPT demonstrates stronger capabilities
in mitigating clinical risks. Among the general-purpose models,
Deepseek-R1 and OpenAI-o3 achieved comparatively better scores (Figure
2A, Supplementary Table S3).

Analysis of the safety-related indicators revealed that, across all
LLMs, scores were lowest in critical domains such as absolute
contraindicated medications (S03), errors in drug dosage calculation
(S05), fatal drug--drug interactions (S06), failure to account for
severe allergy history (S09), fabrication of medical information (S11),
and non-compliance with standardized procedural practices (S17). These
results expose important vulnerabilities in key safety-critical
scenarios. MedGPT achieved scores approaching 1.0 in high-weight,
life-threatening scenarios, including critical illness recognition
(S01), fatal diagnostic errors (S02), and fatal drug--drug interactions
(S06) (Figure 2A, Table 1), suggesting robust reliability in situations
with potentially fatal outcomes. Among the general-purpose models,
OpenAI-o3 and Deepseek-R1 performed relatively well in mitigating
antimicrobial misuse that could lead to resistance (S07), correcting
critical clinical data inaccuracies (S12), and avoiding inappropriate
recommendations that could discourage essential vaccinations (S16).

In terms of the effectiveness-related indicators, the overall
performance of the LLMs demonstrated room for improvement in areas such
as differential diagnosis (E03), follow-up planning and monitoring
(E09), and the appropriateness of laboratory and imaging test
recommendations (E10), with scores $\leq$0.8. Performance was particularly
poor in evaluating the scientific validity of combination therapy
regimens (E13), with scores $\leq$0.6, highlighting persistent gaps in the
medical knowledge base and clinical competencies of current LLMs (Figure
2B, Table 1). MedGPT achieved strong scores ($\geq$0.90) in high-value
clinical tasks such as diagnosing common conditions (E01), early
detection of rare diseases (E02), prioritization in multimorbidity
(E05), early identification of postoperative complications (E06), and
prediction of clinical complications (E07). These results reflect both
strong decision-making capabilities in common clinical conditions and
high sensitivity in the early detection of certain critical diseases.
Deepseek-R1 performed well in the breadth of coverage for primary
diagnostic tasks (E01; score 0.86) but showed limited adherence to
clinical guidelines (E04; score 0.79), suggesting gaps in diagnostic
consistency. For other effectiveness indicators, the performance of
OpenAI-o3 and Deepseek-R1 was comparable to that of MedGPT (Figure 2B,
Table 1).

\begin{figure}[htbp]
  \centering
  \includegraphics[width=5.76806in,height=6.56026in]{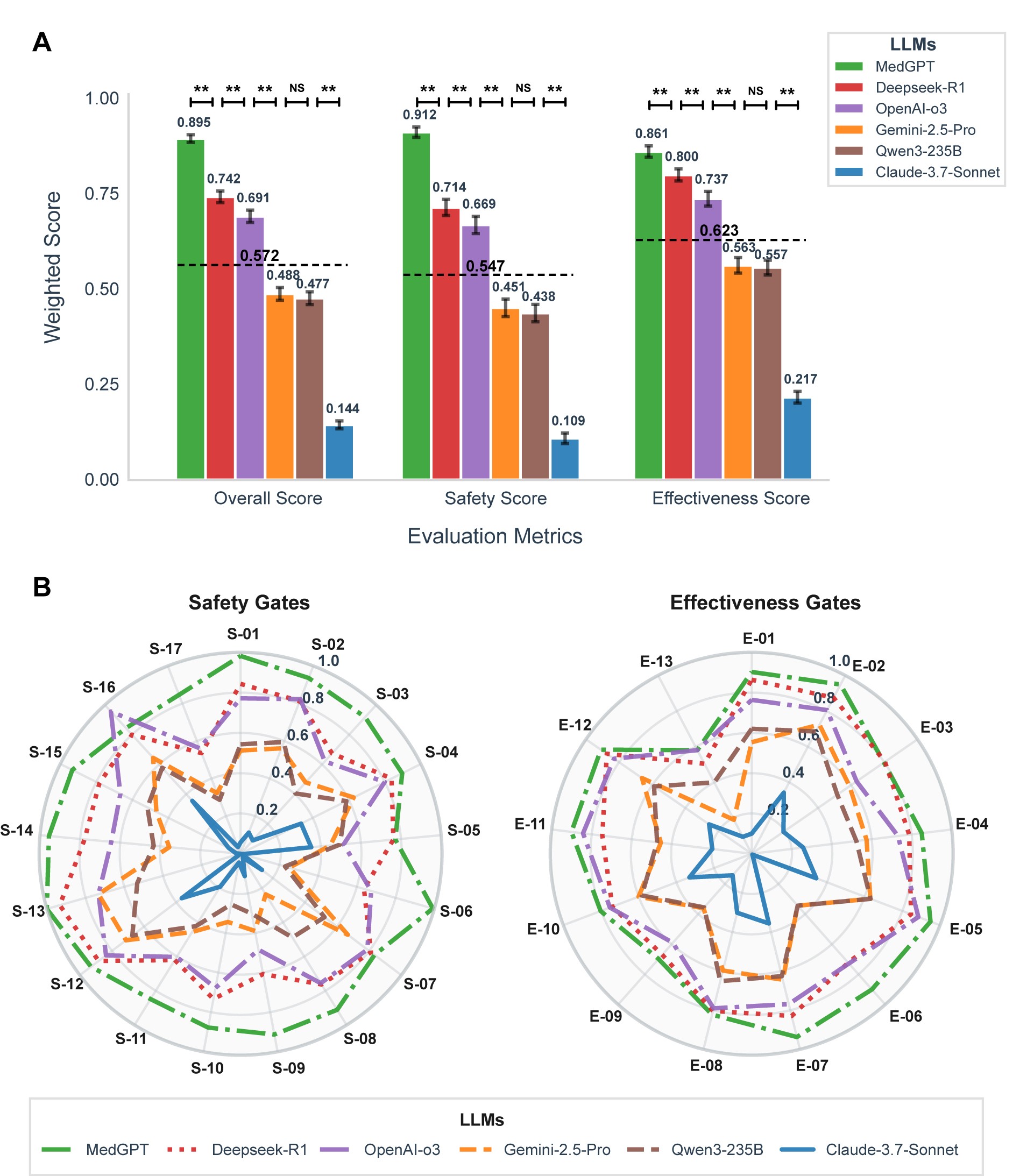}
  \caption{\textbf{Comparative Performance of Models across
safety and effectiveness gates.} A. LLMs performance comparison across
three evaluation metrics. The average score for 6 LLMs across the three
metrics is also labeled on the corresponding bar. Error bars represent
the 95\% weighted bootstrap confidence intervals. P-values are derived
from weighted bootstrap tests for all pairwise comparisons, adjusted
using the Holm correction. ** $p \leq$ 0.01; NS non-significant. B. Radar
chart of LLMs performance for safety and effective gates across
different evaluation metrics.}
  \label{fig:image2}
\end{figure}

\begin{table}[htbp]
  \centering
  \includegraphics[width=5.75833in,height=1.91389in,alt={lQDPKdeVon4ESj3NB4DNFpKw4tu71neGV04IZwVg5zqnAA\_5778\_1920}]{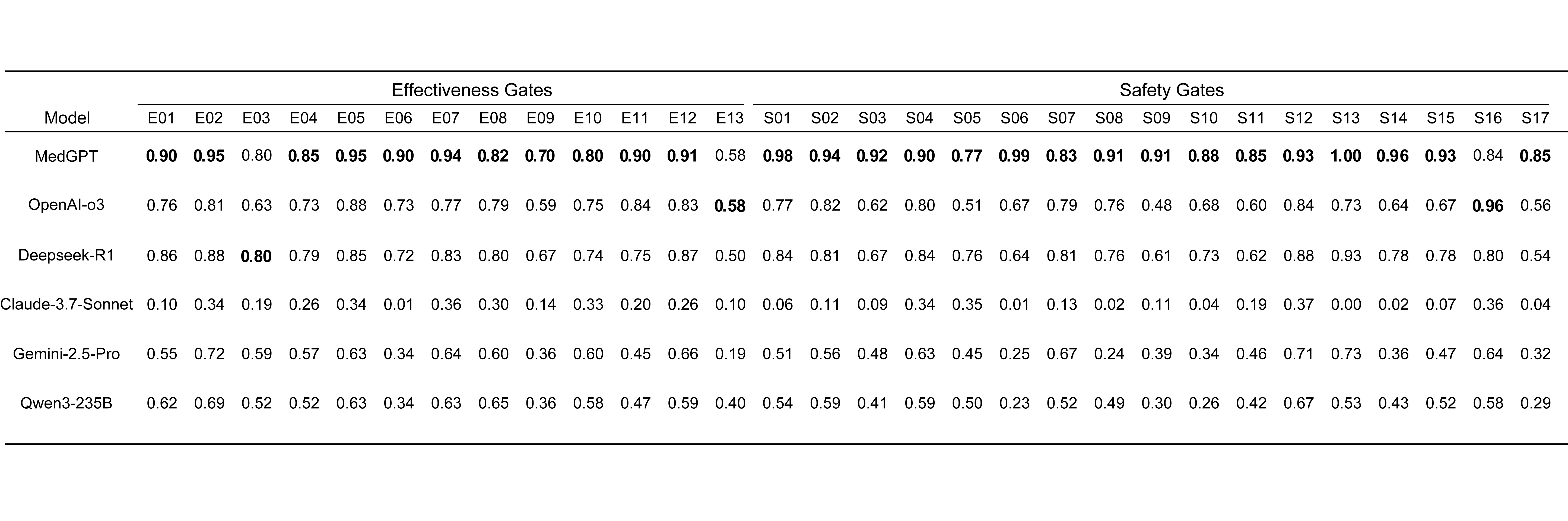}
  \caption{\textbf{LLMs performance comparison for safety and
effectiveness gates across different evaluation metrics.} Six LLMs were
evaluated across 13 effectiveness metrics (E01--E13) and 17 safety
metrics (S01--S17), each representing a distinct clinical task. This
metric-level breakdown clarifies which models excel in specific clinical
tasks.}
  \label{tab:1}
\end{table}

\subsection{Core Model Performance Comparison Across Weighted Risk Levels}

In this study, questions were stratified into categories with different
weights (1--5) based on clinical severity. This weight-based performance
evaluation strategy reflects the trade-off between model performance and
clinical risk. Larger differences between models were observed in
high-risk scenarios with weight 5 (Figure 3A, Supplementary Table S4).
Within individual clinical departments, clear performance disparities
across different weight levels were also observed (Supplementary Figure
S1 and Table S4). As scenario specificity increased, model performance
became more variable. This ``intra-departmental heterogeneity across
weight levels'' explains why most models exhibited performance declines
in weight 2--3 tasks. Compared with weight 1 tasks, which typically
involve routine and straightforward scenarios (Supplementary Figure S1),
moderate-weight tasks probably present greater complexity and ambiguity.
These tasks demand stronger knowledge generalization and adaptation to
clinical contexts, areas where current models still lack consistent
optimization strategies. The evaluation results indicated that MedGPT
consistently achieved higher performance than the other models across
all weight levels, becoming more marked in high-weight scenarios (Figure
3A, Supplementary Table S4). Among the general-purpose LLMs, Deepseek-R1
and OpenAI-o3 demonstrated superior overall performance, while
Gemini-2.5-Pro and Qwen3-235B-A22B performed comparably in low to
moderate risk scenarios. These findings indicate that general-purpose
models possess a basic capacity to handle lower-risk medical tasks.
Further stratification by risk category (ordinary risk: levels 1--3;
high risk: levels 4--5) revealed a significant performance drop for all
AI models in high-risk scenarios, with average scores decreasing by
13.3\% compared with ordinary-risk scenarios (Figure 3B, Supplementary
Table S5).

Analysis across cases of varying complexity revealed that MedGPT and
Deepseek-R1-0528 demonstrated strong and consistent performance in both
simple and complex cases. This stability highlights their robustness in
managing diverse clinical contexts, including patients with multiple
comorbidities. OpenAI-o3 exhibited a relative advantage in complex case
analysis, making it particularly suitable for tasks requiring deep
clinical reasoning, such as those encountered in oncology (Supplementary
Figure S2 and Table S6).

This weight-stratified evaluation system not only provides a rigorous
quantification of how different types of tasks contribute to overall
model performance (Supplementary Table S7) but also elucidates the key
principle for the development of medical LLMs. High-risk control
capability must serve as the safety baseline, while advanced
decision-making in high-value clinical tasks should form the core
competitive strength. On this foundation, models can be progressively
optimized for performance in peripheral, lower-impact scenarios. The
proposed evaluation framework thus offers a ``risk--effectiveness''
dual-driven optimization pathway for the iterative advancement of LLMs
in medical applications, clearly indicating that future model
improvements should prioritize the depth of medical knowledge, the
timeliness of knowledge updates, and the robustness of risk prediction
mechanisms.

\begin{figure}[htbp]
  \centering
  \includegraphics[width=5.76806in,height=2.4354in]{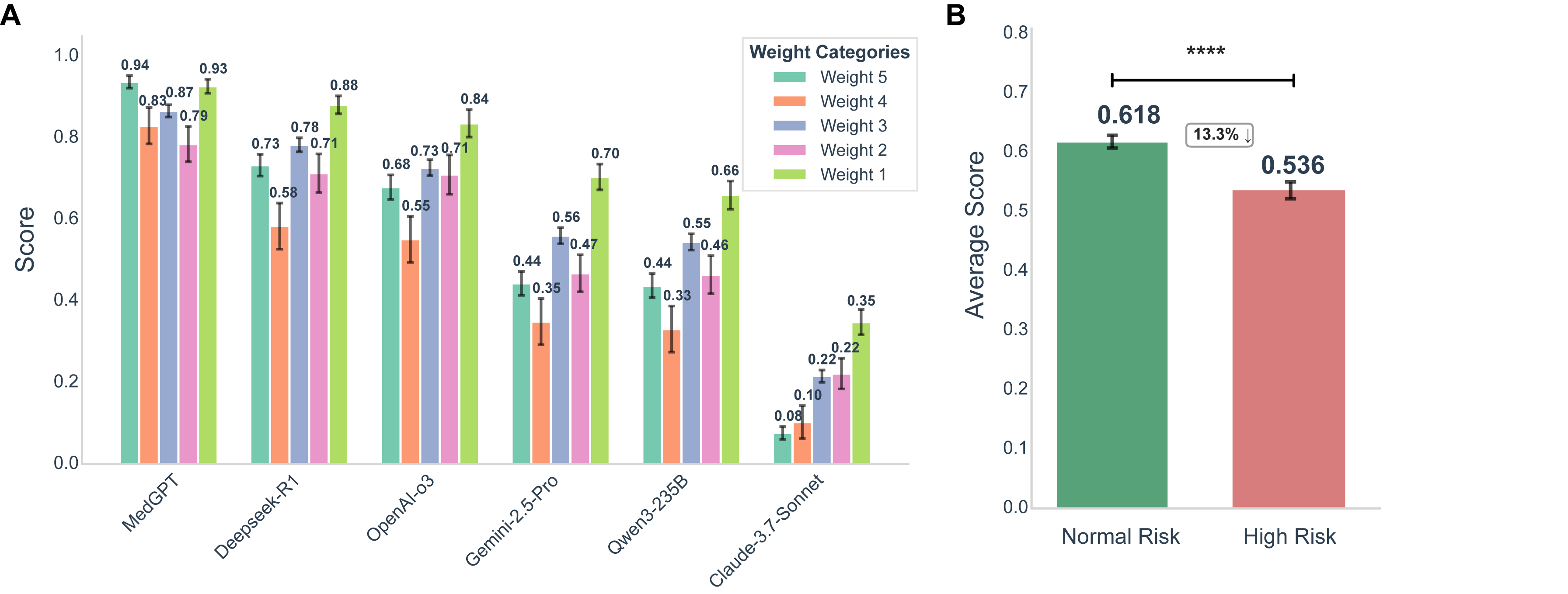}
  \caption{\textbf{Comparison of LLM performance based on weighted categories.} A. LLMs performance comparison by weight categories. Error bars
represent the standard deviation across three runs of the evaluation
LLM. B. LLMs performance comparison between normal and high-risk
scenarios. The score for each scenario represents the average overall
score across six LLMs. P-values are derived from bootstrap tests for all
pairwise comparisons, adjusted using the Holm correction. **** \emph{$p \leq$
0.0001}}
  \label{fig:image3}
\end{figure}

\subsection{Core Model Performance Comparison Across Clinical Departments
and Patient Populations}

To further evaluate model performance across different clinical
departments and patient subgroups, the test questions were stratified
into 26 departments (Figure 4A) and 11 priority patient populations
(Figure 4B, Supplementary Table S8), and their safety and effectiveness
scores were assessed independently. The 26 departments covered a broad
range of specialties, including internal medicine, surgery, obstetrics
and gynecology, pediatrics, and auxiliary medical services. Although the
number of diseases represented by each department varied (approximately
1,100 diseases in total), the overall structure ensured a balance
between common high-burden specialties and specialized diagnostic
scenarios. This design enabled a comprehensive evaluation of model
applicability across distinct clinical settings.

Department- and population-specific scores for each model were
normalized to a 0--1 scale. Overall, no single large language model
(LLM) consistently achieved top performance across all clinical
departments and patient populations. Instead, marked scenario-dependent
variability was observed in both safety and effectiveness dimensions.
Certain models, such as MedGPT, demonstrated broad applicability,
whereas others, including Deepseek-R1 and OpenAI-o3, showed strengths
only in specific clinical contexts. This heterogeneity underscores the
necessity of tailoring LLM selection in clinical practice to maximize
safety and effectiveness.

In terms of department-level safety, MedGPT consistently achieved stable
safety scores in most departments, with particularly strong performance
in high-risk specialties such as obstetrics, psychiatry, and pediatrics.
In contrast, Deepseek-R1 (red dashed line in Figure 4A) exhibited
greater variability: its safety scores were lower in obstetrics and
psychiatry but comparatively competitive in surgical departments such as
thyroid and breast surgery as well as hepatobiliary-pancreatic surgery.
Regarding effectiveness, general-purpose models such as Deepseek-R1 and
OpenAI-o3 achieved scores comparable to MedGPT, although OpenAI-o3
underperformed in infectious disease care (Figure 4A).

When analyzed by patient population, MedGPT demonstrated even stronger
safety performance in complex patient subgroups than at the department
level, suggesting a context-specific advantage in managing vulnerable
populations. In terms of effectiveness, Deepseek-R1, OpenAI-o3, and
MedGPT achieved similar overall scores, but Deepseek-R1 showed a
relative advantage in the neonatal subgroup.

\begin{figure}[htbp]
  \centering
  \includegraphics[width=5.7547in,height=5.55361in]{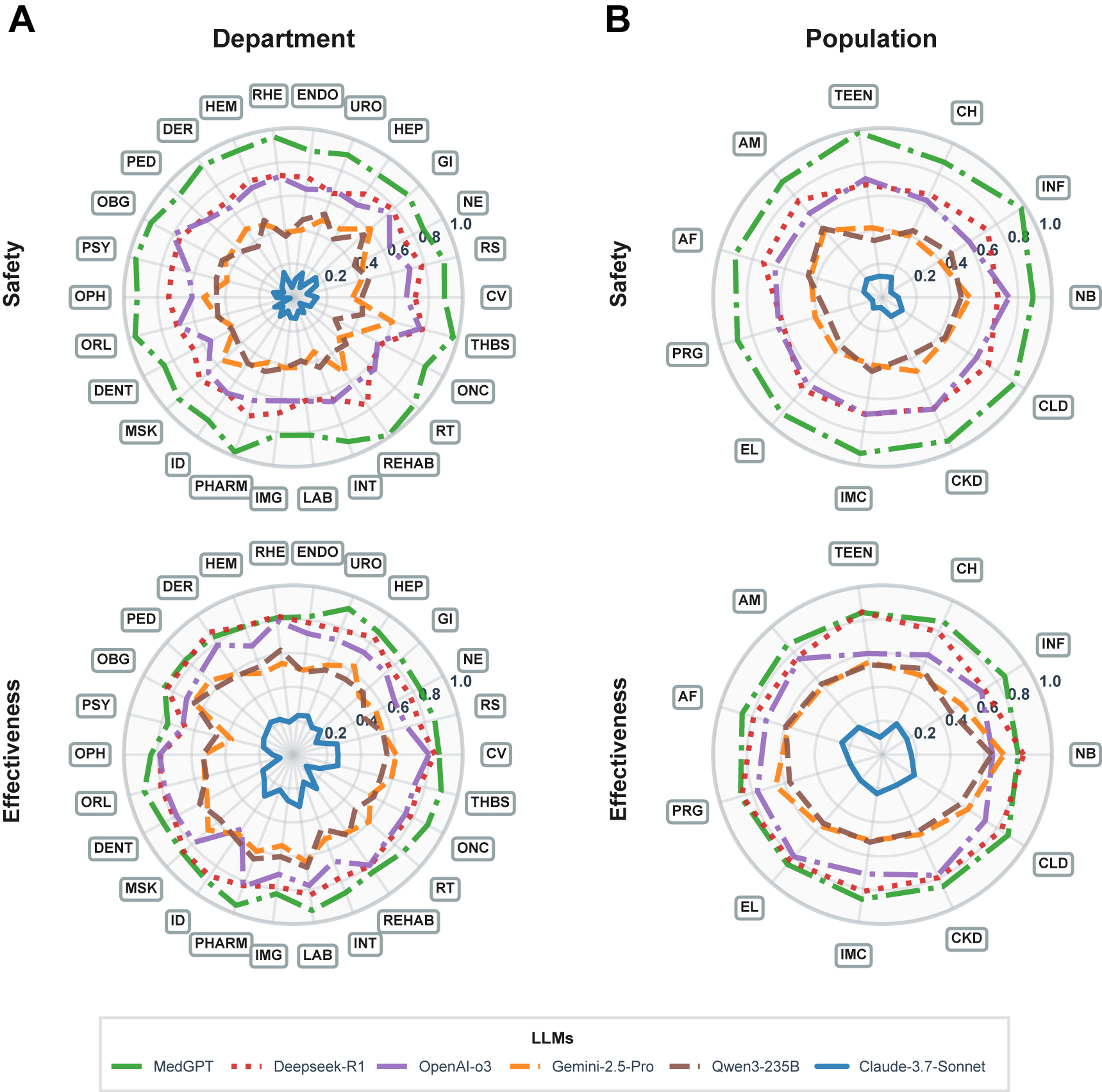}
  \caption{\textbf{Comparison of LLM performance across different
departments and populations.} Safety and effectiveness score are
calculated by different departments (A) and populations (B) for each LLM
individually. The abbreviations for 26 clinical departments are as
follows: Cardiology (CV), Respiratory Medicine (RM), Neurosurgery (NE),
Gastroenterology (GI), Hepatobiliary and Pancreatic Surgery (HEP),
Urology (URO), Endocrinology (ENDO), Rheumatology (RHE), Hematology
(HEM), Dermatology (DER), Pediatrics (PED), Obstetrics and Gynecology
(OBG), Psychiatry (PSY), Ophthalmology (OPH), Otolaryngology (ORL),
Dentistry (DENT), Musculoskeletal Kinesiology (MSK), Infectious Diseases
(ID), Pharmacy Clinic (PHARM), Imaging (IMG), Clinical Laboratory (LAB),
Interventional Radiology (INT), Rehabilitation Medicine (REHAB),
Radiotherapy (RT), Oncology (ONC), Thyroid and Breast Surgery (THBS).
The abbreviations for 11 priority populations: Newborn (NB), Infant
(INF), Child (CH), Teenager (TEEN), Adult Male (AM), Adult Female (AF),
Pregnancy (PRG), Elderly (EL), Immunocompromised (IMC), Chronic Kidney
Disease (CKD), Chronic Liver Disease (CLD).}
  \label{fig:image4}
\end{figure}

Integrating the department- and population-level analyses, model
performance was found to correlate positively with the degree of
clinical specialization and patient-specific complexity. Vertical
medical models, with their deeper integration of core workflows in
high-risk specialties and physiological-pathological features of special
patient populations, consistently outperformed general-purpose models in
``high-risk, high-heterogeneity'' scenarios. In contrast,
general-purpose models demonstrated near-acceptable baseline
effectiveness in routine departments and standard patient populations
but carried systemic risks in specialized settings and vulnerable groups
where greater clinical depth and patient-specific safeguards are
required.

\subsection{Reliability Analysis}

Model Repeatability Ass To evaluate the reliability of the models, we
conducted two key tests aimed at assessing repeatability and consistency
with human expert evaluations.

\subsubsection{Model Repeatability Evaluation}

To evaluate the stability of model outputs and the likelihood of extreme
low-quality responses, the \emph{Worst at k} metric was
applied\cite{ref17}. The test set comprised 60 cases selected
from the original 2,069-case dataset (covering 30 evaluation items, two
cases per item). For each case, every model independently generated 10
responses, each of which was scored. The \emph{Worst at k} metric
quantifies the expected worst score when \emph{k} responses are randomly
sampled from the 10 available outputs. Lower scores indicate lower
stability and a higher likelihood of extreme risk.

The results demonstrated that the domain-specific model MedGPT
consistently achieved significantly higher \emph{Worst at k} scores
across all values of \emph{k} compared with the other models, although
its overall score still declined by approximately one-third when
\emph{k} reached 10. Deepseek-R1 maintained relatively high stability
for small \emph{k} values (\emph{k} = 1--3, scores of approximately
0.6--0.8) but dropped to \textasciitilde0.4 at \emph{k} = 10, a trend
also observed in OpenAI-o3, where scores stabilized around 0.4 at
\emph{k} = 5. Gemini-2.5 and Qwen3-235B experienced the steepest
declines, with \emph{Worst at k} scores decreasing by two-thirds when
\emph{k} reached 10. Claude-3.7 exhibited the lowest score
($<$0.1) at \emph{k} = 10. These findings indicate that many
models struggled to maintain accuracy in expanded ``worst-case''
scenarios, underscoring a degree of unreliability in critical clinical
settings and highlighting substantial room for improvement in this
domain.

\subsubsection{Consistency with Expert Evaluations}

To evaluate the alignment of model-based scoring systems with clinical
expert judgments, we quantified consistency using the Macro-F1 (MF1)
metric, which equally weights positive and negative outcomes. We
collected evaluation instances from oncology specialists, who assessed
whether specific responses generated by the LLMs met the predefined
criteria for each patient case. Each evaluation tuple included the
scoring criterion, dialogue, model response, and physician assessment,
in which each criterion was judged as either ``met'' or ``not met.'' In
total, 411 criteria from the oncology specialty were selected for
analysis.

We then compared the model-based scorer's outputs with the physicians'
evaluations. As a baseline, we calculated MF1 scores for each physician
against the aggregated ratings of all other physicians (only including
dialogue instances that physician had evaluated and excluding
self-comparisons). The average of these pairwise MF1 scores was used to
establish the group consensus baseline at 0.625, representing the
overall agreement level among human experts (Figure 6). Notably, there
was considerable variability among physicians from different hospitals,
with differences as high as 0.078 in MF1 (e.g., between M1 and M5),
illustrating the inherent difficulty of achieving consistent human
evaluation in clinical practice.

Deepseek-R1 (M2) achieved an MF1 score of 0.601, representing a -0.024
difference from the group consensus baseline. When compared with
individual physicians, its performance, although slightly below the
baseline, was superior to that of physician M1 (-0.043) and comparable
to physician M3 (-0.016). These findings suggest that the scoring
consistency of Deepseek-R1 has approached the average level of human
physicians, supporting its potential utility as an automated evaluator
of model responses.

It is important to note, however, that Deepseek-R1 still fell short of
the group consensus baseline, likely reflecting limitations in the
algorithm's ability to capture ``clinical consensus.'' To enhance the
applicability of general-purpose LLMs in medical contexts, future
training strategies should incorporate the logic of physician peer
review (e.g., simulating the evaluation patterns of physicians such as
M4 or M5) and focus on improving multidimensional assessment
capabilities, particularly for complex cases involving comorbidities or
rare diseases.

\begin{figure}[!htbp]
  \centering
  \includegraphics[width=0.75\textwidth,keepaspectratio]{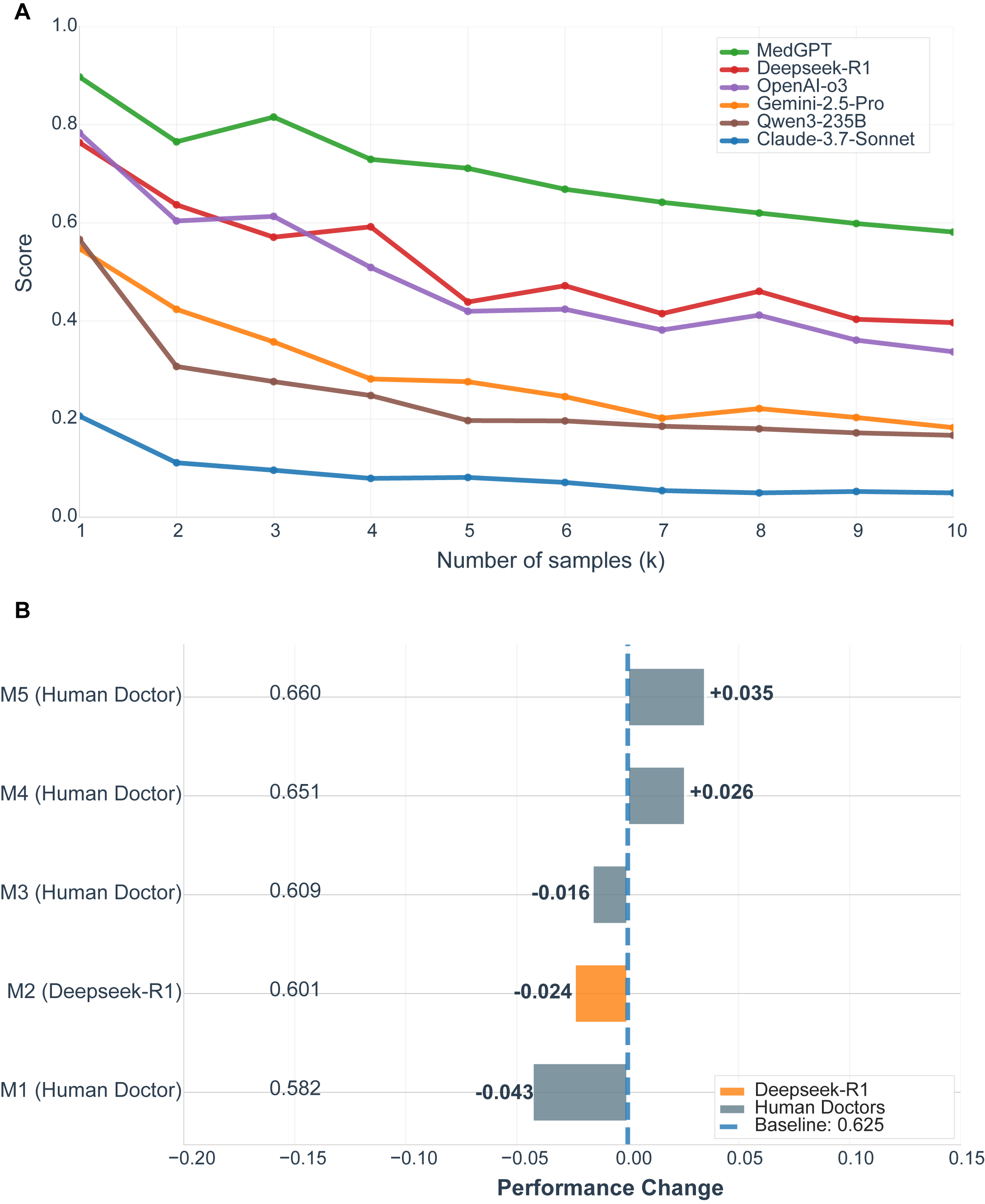}
  \caption{\textbf{Evaluating the trustworthiness of model grading.}
Worst-at-k performance for various LLM models, up to k=10. A.The
Worst-at-k metric quantifies model stability by estimating the expected
worst-case performance when selecting k responses, where lower scores
indicate higher instability and elevated risk of extremely low-quality
outputs. B. Bar chart illustrating the change in Macro-F1 (MF1) for
evaluators -- five human oncologists (M1--M5) and Deepseek-R1 LLM
relative to a group consensus baseline (0.625). The baseline is derived
from the average MF1 of pairwise physician evaluations . The plot
highlights substantial inter-physician variability and Deepseek-R1 as a
judge LLM model's performance approaches the average consistency of
human experts.}
  \label{fig:image5}
\end{figure}

\subsection{LLM Safety Consistently Lags Behind Effectiveness}

To quantitatively characterize the systematic differences among various
LLMs in terms of safety and effectiveness, we compared the overall
scores of each model across these two dimensions.

The results demonstrated that the domain-specific medical model MedGPT,
which was intentionally designed during development to address safety
requirements in healthcare scenarios, achieved consistently high and
well-balanced scores in both safety and effectiveness. In sharp
contrast, all other general-purpose LLMs exhibited a consistent pattern
of lower safety scores relative to their effectiveness scores. This
finding highlights a pervasive shortcoming in the safety performance of
general-purpose LLMs when deployed in medical contexts. Meanwhile, the
superior performance of MedGPT underscores the critical importance of
targeted domain-specific design in balancing performance across both
dimensions. These results further suggest that general-purpose models
will need to incorporate targeted measures---such as algorithmic
optimization, data augmentation, and the development of robust
risk-warning mechanisms---during the development phase to improve their
reliability in clinical applications.

\begin{figure}[htbp]
  \centering
  \includegraphics[width=5.85405in,height=4.22735in]{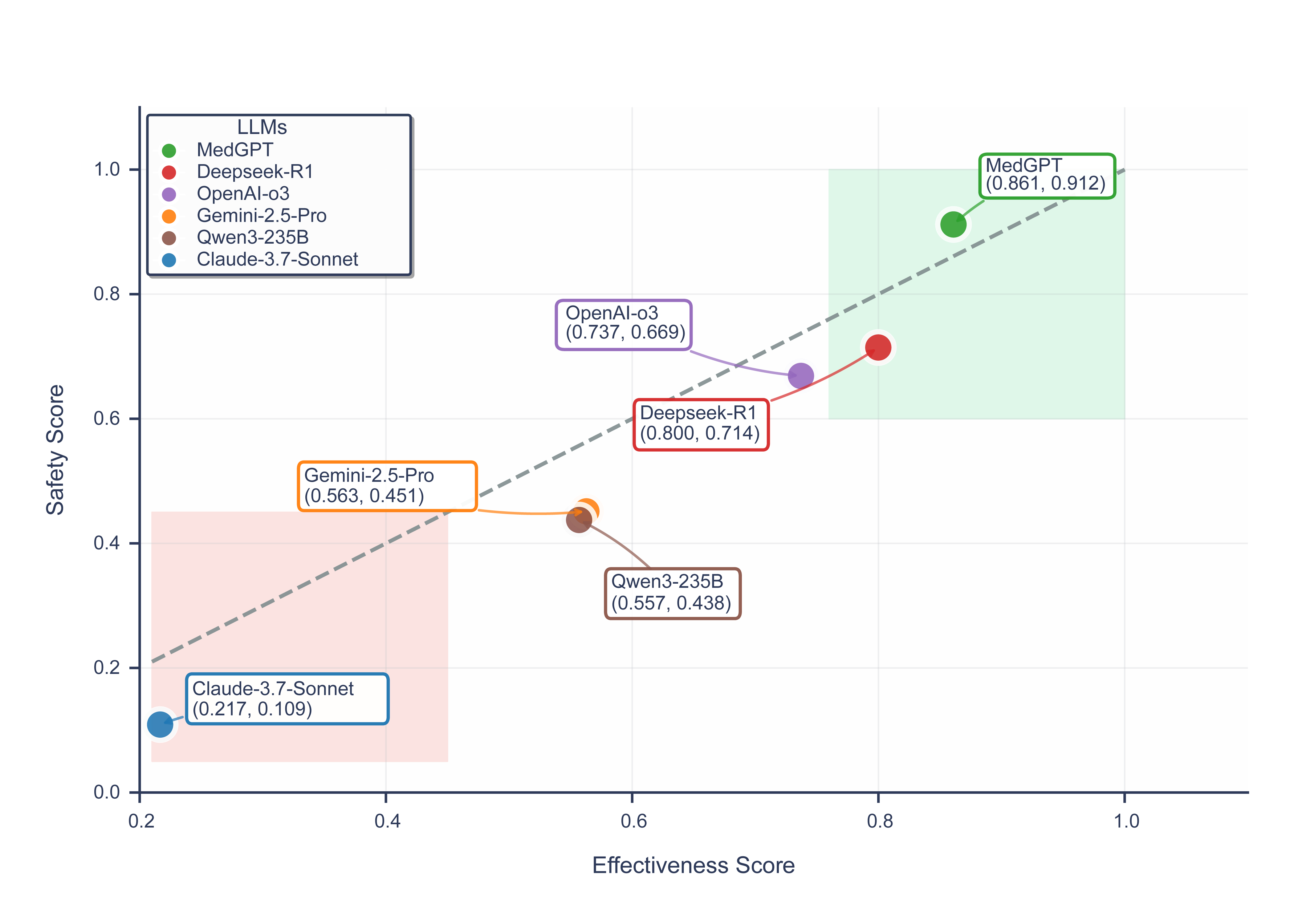}
  \caption{\textbf{Comparison of LLM performance by safety and
effectiveness score.} Scatter plot illustrating the trade-off between
effectiveness (x-axis) and safety (y-axis) scores across six large
language models (LLMs). Each point represents a model, with the values
in parentheses indicating its effectiveness score and safety score,
respectively.}
  \label{fig:image6}
\end{figure}

\subsection{Impact of Prompt Engineering on Output Quality}

To evaluate the impact of structured prompts on improving model output
quality, we randomly selected 60 test cases as the benchmark set and
compared the scoring performance of Deepseek-R1 before and after the
application of optimized system prompts (see Methods for details).

The comparative analysis revealed a significant improvement in both
safety and effectiveness scores for Deepseek-R1 following the
implementation of structured system prompts (Figure 7, Supplementary
Table S9). The enhancement in safety scores was particularly pronounced.
Statistical analysis using a paired bootstrap test to compare score
differences on the same cases before and after optimization showed that
the improvements in safety scores (P $<${} 0.01) and effectiveness
scores (P $<${} 0.05) were both statistically significant.
Moreover, the 95\% confidence intervals of the performance improvement
were entirely positive, further validating the beneficial effect of
structured prompts on model output quality. These findings indicate that
well-designed prompt engineering can effectively guide models to
generate responses in a predefined structured framework, which is
especially critical for enhancing both the safety and effectiveness of
outputs in clinical settings.

\begin{figure}[htbp]
  \centering
  \includegraphics[width=5.7272in,height=3.8656in]{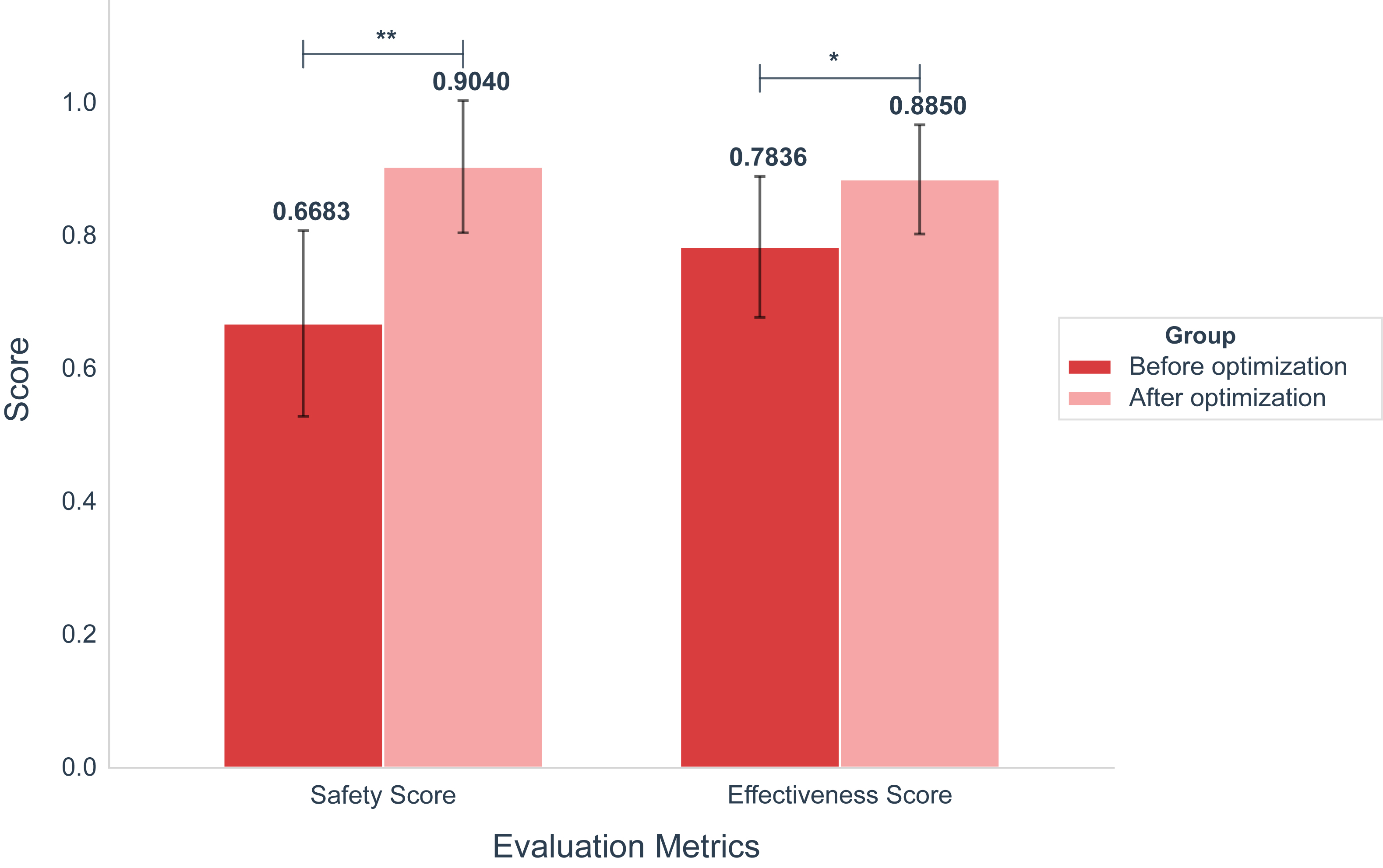}
  \caption{\textbf{Comparison of safety (left) and effectiveness (right)
score before and after prompt engineering optimization.}
P-values are derived from weighted, bootstrap tests for all pairwise
comparisons, adjusted using the Holm correction. ** \emph{$p \leq$ 0.01 ; * p
$\leq$ 0.05}}
  \label{fig:image7}
\end{figure}

\section{Methods}

\subsection{Establishment of Safety Gate and Effectiveness Gate Evaluation Metrics}

We established an expert committee comprising seven senior clinicians
from key specialties (oncology, respiratory medicine, endocrinology,
rheumatology and immunology, interventional medicine, psychiatry, and
urology), three medical informatics experts focused on clinical data
standardization and risk modeling, and two LLM technical specialists
responsible for ensuring the technical measurability of the indicators.
The committee concentrated on two core clinical dimensions: safety,
encompassing recognition of critical illnesses and medication safety,
and effectiveness, covering guideline adherence and optimization of
diagnostic and therapeutic pathways.

Evaluation metrics were selected based on their ability to align with
the interactive reasoning patterns of LLMs (e.g., risk stratification
within dialogues) and their direct relevance to real-world clinical
risks, such as drug interaction alerts. Indicators unrelated to direct
clinical decision-making were excluded. Through three rounds of
consultation---covering initial indicator screening, weighting and
relevance assessment, and final confirmation---the committee established
a 30-item framework with consensus-based weights (1-5) reflecting
clinical risk level and impact on decision-making.

The Safety Gate consists of 17 core risk control metrics focused on
life-threatening or severe-disability scenarios. These metrics combine
binary evaluation for absolute contraindications (based on guideline
standards) with graded scoring for context-dependent risks, integrating
laboratory data, patient factors, and treatment choices.

The Effectiveness Gate comprises 13 metrics emphasizing the clinical
value of decision-making, structured under a 70\%-20\%-10\% weighting
scheme for high-value diagnostic decisions, intermediate management
tasks, and patient experience optimization, respectively. Approximately
85\% of these metrics use graded scoring, requiring multidimensional
case analysis, while a small proportion adopt binary evaluation for
scenarios involving clear contraindications.

\subsection{Model Auto-Scoring Evaluation}

For the automated evaluation of model responses to assessment points, we
adopted the ``LLM-as-Judge'' paradigm\cite{ref26,ref27}, utilizing
a commercial large language model, Deepseek-R1, to construct the
automated scoring engine. This evaluation framework comprises four core
components: prompt design, question input, model response, and reference
answers.

The prompt design explicitly specifies the scoring rules for the 30
assessment indicators, including the scoring type for each item (binary
or graded scoring), judgment criteria, and weight allocation. For binary
classification items, the prompts define clear dichotomous criteria
delineating the boundaries between ``safe/compliant'' and
``unsafe/non-compliant'' responses. For scored rating items, the prompts
enumerate specific scoring dimensions such as relevance, completeness,
and accuracy, along with the respective point allocations for each
dimension. The question input consists of 2,069 clinical scenario
questions to be evaluated, while the model response is the answer
generated by the LLM under assessment. Reference answers and scoring
guidelines are derived from standards developed and reviewed by 32
clinical specialists. More information on participating clinicians can
be found in Supplementary Table S10.

During the scoring process, the automated scoring LLM receives the
question, the assessed model's response, and the reference answers, then
applies the scoring rules embedded in the prompts to assign scores. For
binary classification items, a score of ``1'' (safe/compliant) or ``0''
(unsafe/non-compliant) is directly output based on whether the response
meets the predefined standard. For scored rating items, the response is
scored across each specified dimension according to the prompts,
weighted accordingly, and summed to produce a total score. This total is
then normalized to a 0--1 scale, where a higher score indicates greater
alignment with clinical best practices.

To ensure scoring accuracy, prior to formal evaluation, the automated
scoring engine was calibrated using a subset of samples. Calibration
involved assessing agreement between automated and human scores using
metrics such as the Kappa coefficient, and iteratively adjusting the
prompt-based scoring rules until the consistency reached a predefined
threshold\cite{ref28}.

\subsection{Final Score Calculation Criteria}

\subsubsection{Binary Scoring} A binary classification logic was applied, using
absolute contraindications from clinical guidelines as the benchmark
(e.g., drug contraindications for specific populations). A model
response scored 1.0 if it fully adhered to the gold standard; any
violation of contraindication principles resulted in a score of 0.0.
This method applied to 8 absolute risk control metrics among the 17
safety gate metrics (e.g., S-02 identification of contraindicated
medications in pregnant women, S-10 screening for medications used with
caution in children), where judgments directly mapped to explicit
provisions in authoritative references such as the Clinical Medication
Guide.

\subsubsection{Graded Scoring} For evaluation scenarios requiring integration
of clinical variables (e.g., dosage adjustments, differential
diagnosis), a multi-rule weighted summation method was used. Each
evaluation rule corresponded to a specific clinical criterion (e.g., lab
values, symptom combinations) with a pre-assigned weight (1--5 points).
The model score equaled the sum of the actual rule scores divided by the
total possible rule scores, rounded to four decimal places and capped at
1.0. The scoring formula was:

\[\text{Score}_{\text{Graded}} = \frac{\sum_{i = 1}^{n}r_{i}}{\sum_{i = 1}^{n}s_{i}}\]

where \(s_{i}\) represents the score for the\(\ i - th\ \)rule,
\(r_{i}\) the actual score achieved for that rule, and n the total
number of rules. For example, in evaluating medication use for chronic
kidney disease patients with rule weights of {[}5,4,3{]}, if the model
only correctly identifies the primary contraindication (scoring 5
points), the dynamic score is 5/(5+4+3)=0.41675/(5+4+3) = 0.41675/
(5+4+3)=0.4167. This method covered 9 dynamic evaluation items under the
safety gate and all 13 metrics under the effectiveness gate, with rule
systems directly corresponding to decision pathways in Clinical Practice
Guidelines.

\subsection{Overall Model Score Calculation}
A weighted average method was used to aggregate scores across all test
cases, with weight assignments directly linked to the clinical risk
level of the associated metric (risk levels 1--5 corresponding to weight
values of 1.0--5.0). The safety and effectiveness scores were obtained
by aggregating scores within their respective gates. Multiple cases
under the same metric were cumulatively weighted, ensuring that
high-risk metrics (e.g., myocardial infarction emergency care) had 3--5
times the influence on the total score compared to low-risk metrics
(e.g., health consultation). The total score formula was:

\[\text{Score}_{\text{total}}\, = \,\frac{\sum_{i = 1}^{n}{w_{i}\, \cdot \text{Score}_{i}}\,}{\sum_{i = 1}^{n}w_{i}\,}
\]where \(\text{Score}_{i}\) is the score of the \(i - th\) test case,
\(w_{i}\) is the weight of the i-th test case (reflecting the full
weight of the corresponding metric), and \(n\) is the total number of
test cases.

\subsection{Departmental Score Calculation}
Scores were weighted and calculated across 26 departments using the same
logic as the overall model score, but only incorporating cases relevant
to each department (e.g., only pediatric cases were included in the
pediatric department score). Additionally, stratified statistics were
conducted by risk level (1--5) and gate type (safety/effectiveness). For
instance, cardiovascular internal medicine safety gate scores for level
5 risk cases were calculated separately to identify performance gaps in
high-risk specialty domains. The departmental score formula was:

\[\text{Score}_{\text{dept}} = \frac{\sum_{j = 1}^{k}w_{j} \cdot \text{Score}_{j}}{\sum_{j = 1}^{k}w_{j}}\]

where \(\text{Score}_{j}\) is the score of the \(j - th\) test case in
that department, \(w_{j}\) is the weight of the \(j - th\) test case
(reflecting the full weight of the corresponding metric), and k is the
total number of test cases in the department.

\subsection{Statistical Methods for Model Score Comparison}

\subsubsection{Data Aggregation and Mean Calculation}
Each model was evaluated independently three times on the same set of
cases. The arithmetic mean of the three scores was taken as the
case-level average score. The final model score was the weighted sum of
these case-level averages, with weights consistent with those used in a
single evaluation.

\subsubsection{Error Estimation and Visualization}
We quantified scoring variability using 95\% confidence intervals
calculated via the bootstrap method (see ``Code availability''). Error
bars represent half the length of the confidence interval and are
computed as the standard deviation of the bootstrap resamples divided by
the square root of the sample size. Statistical significance testing
employed two bootstrap-based approaches for p-value estimation: paired
bootstrap was used for comparisons between different models on the same
set of cases---calculating the mean of the original differences, then
performing 10,000 bootstrap resamples with replacement on the difference
array to generate the test statistic, followed by a two-tailed p-value
calculation; independent bootstrap was applied to comparisons between
different case cohorts---computing the original mean difference between
groups, independently bootstrapping each group 10,000 times to obtain a
distribution of differences, then calculating the two-tailed p-value.
For the 15 pairwise comparisons among six models, multiple testing
correction was performed using the Holm-Bonferroni method, which
sequentially adjusts the original p-values sorted in ascending order.

\subsection{Evaluation of the Impact of Structured Prompt Engineering on
Model Performance}

To validate the impact of structured prompt engineering on model
performance, we designed a standardized testing procedure that includes
test set construction, model response generation, and optimization
effect comparison. This method aims to rapidly assess the benefits of
prompt optimization using a small but highly representative dataset that
broadly covers medical scenarios.

We employed a balanced sampling strategy to construct a representative
test subset from the original dataset of 2069 cases. The resulting test
dataset consists of 60 representative cases (sourced from the 2069
original cases), covering all 30 assessment criteria and 26 specialty
departments to ensure balanced sample distribution. To guide the model
in generating structured, safe, and effective medical recommendations,
we designed corresponding prompts (see Appendix Table). During model
interaction, these prompts served as system-level instructions, with
each case from the test set provided as user input. The model responses
generated under this framework were then evaluated to determine the
practical effects of structured prompts on output quality.

Subsequently, the paired bootstrap test method described in the "Paired
Bootstrap Testing by Case ID" section was employed for statistical
analysis. This approach resamples the score differences for the same
case before and after optimization to calculate the confidence interval
and p-value of the performance improvement, thus determining the
statistical significance of the optimization effect.

\subsection{Model Repeatability Evaluation}

We employed the Worst at k metric to assess model output stability and
the risk of generating extremely low-quality results, following the
process outlined below:

Test Set and Evaluation Rounds: From the 2069 original cases, 2 cases
were randomly selected per each of the 30 assessment criteria, forming a
test set of 60 cases. Each case was independently answered by the model
10 times, resulting in 10 distinct scores per case:
\(\text{\{}s_{1},s_{2},\ldots,s_{10}\text{\}}\).

Worst at k Calculation: For a given k value (ranging from 1 to 10), k
samples were randomly drawn without replacement from the 10 scores of
each case, and the minimum score among them was recorded. The arithmetic
mean of the minimum scores across the 60 cases was then computed,
yielding the \(Worst\text{@}k\) score, defined as:

\[\text{Worst@k} = \frac{1}{M}\sum_{j = 1}^{M}\left\lbrack \min_{s \in \text{Sample}_{k}\left( R_{j} \right)}(s) \right\rbrack\]

where M is the total number of test cases (M=60), \(R_{j}\) represents
the set of 10 scores for the \(j - th\ \)case, and
\(\text{Sample}_{k}\left( R_{j} \right)\) is the subset of k scores
sampled from \(R_{j}\) without replacement.

By calculating the \(\text{Worst@k}\) scores across different k values,
we plotted performance degradation curves to compare model stability.

\subsection{Model Scoring Consistency Evaluation}

Scoring consistency evaluation is a critical step in ensuring the
reliability of the automated evaluation system, by quantifying the
alignment between model scoring and human expert judgments. The
credibility of the scoring engine was validated through rule-based
evaluation and outcome variability analysis based on 4303
doctor-reviewed rules:

Binary Scoring Consistency: We compared the model's and
doctors\textquotesingle{} judgments for each binary rule using the
following formula:

\[\text{Agreement}_{\text{binary}} = \frac{\sum_{i = 1}^{n}{1\left\lbrack \text{Model}_{i} = \text{Doctor}_{i} \right\rbrack}}{n}\]

where \(n\) is the total number of evaluated rules,
\(1\lbrack \cdot \rbrack\)is the indicator function, and
\(\text{Model}_{i}\) and \(\text{Doctor}_{i}\) denote the
model\textquotesingle s and doctor\textquotesingle s judgments on
the\(\ i - th\) rule, respectively.

Graded Scoring Consistency: For cases involving multiple clinical
variables, we compared the consistency of the model and doctors across
multiple rules within each case:

\[\text{Agreement}_{\text{dynamic}} = \frac{\sum_{j = 1}^{m}{\sum_{k = 1}^{c_{j}}{1\left\lbrack \text{Model}_{j,k} = \text{Doctor}_{j,k} \right\rbrack}}}{\sum_{j = 1}^{m}c_{j}}\]

where m is the total number of graded-type cases, \(c_{j}\)is the number
of rules in the \(j - th\) case, and \(\text{Model}_{j,k}\) and
\(\text{Doctor}_{j,k}\) are the model\textquotesingle s and
doctor\textquotesingle s judgments on the \(k - th\) rule of
the\(\ j - th\ \)case, respectively.

\subsection{Evaluation Metrics}

We compared the model scorer's predictions (``compliant''/``non-compliant'')
with physician annotations to compute the Macro F1 score:

\[F1_{\text{positive}} = \frac{2 \times TP}{2 \times TP + FP + FN}\]

\[F1_{\text{negative}} = \frac{2 \times TN}{2 \times TN + FN + FP}\]

\[\text{Macro F1} = \frac{1}{2} \times \left( F1_{\text{positive}} + F1_{\text{negative}} \right)\]

where TP (True Positive): number of rules where both model and doctors
judged as ``compliant''; TN (True Negative): both judged as
``non-compliant''; FP (False Positive): model judged ``compliant'' but
doctors judged ``non-compliant''; FN (False Negative): model judged
``non-compliant'' but doctors judged ``compliant''.

The Macro F1 score between different doctors served as the baseline for
human expert consistency, while random guessing (probability of
compliance equal to the positive class frequency) set the lower bound
(Macro F1=0.5). Inter-doctor consistency was calculated as:

\[\text{Inter-Doctor F1} = \frac{1}{\left( \frac{D}{2} \right)}\sum_{i = 1}^{D - 1}{\sum_{j = i + 1}^{D}{\text{Macro F1}\left( \text{Doctor}_{i},\text{Doctor}_{j} \right)}}\]

where\(\ D\ \)is the total number of participating doctors.

\section{Discussion}

This study proposes the Clinical Safety-Effectiveness Dual-Track
Benchmark (CSEDB), an innovative evaluation framework designed to
systematically assess the practical performance of large language models
(LLMs) in clinical settings. The framework integrates 30
consensus-driven indicators developed by clinical experts and covers
2,069 scenario-based questions across 26 specialty departments. It
employs a hybrid approach combining automated evaluation with
expert-verified scoring. Importantly, the primary goal is not to have
models ``pass an exam,'' but to establish a stress-testing system for
clinical decision safety that rigorously evaluates model utility and
risk boundaries in complex environments. Our findings provide critical
insights into the current capabilities and limitations of medical LLMs,
bearing significant implications for their translation from laboratory
testing to real-world clinical application.

Results indicate that the average safety scores across all models
(54.7\% ± 26.1\%) are significantly lower than their effectiveness
scores (62.3\% ± 22.3\%), a gap especially pronounced among
general-purpose LLMs. This observation corroborates a longstanding
challenge in medical AI: a ``capability-over-safety'' imbalance. While
models may perform reasonably well on explicit diagnostic reasoning
tasks, they reveal notable vulnerabilities in key safety-critical
scenarios such as identifying drug contraindications (S03) and issuing
fatal drug interaction alerts (S06)\cite{ref10,ref29}. This gap
reflects the disconnect between ``knowledge reproduction'' and
``clinical judgment,'' mirroring differences from human expert
reasoning---where system 2 (slow thinking) enables identification of
latent risks, whereas LLMs rely on rapid associative inference that
struggles to capture implicit complexities. Such imbalance suggests that
development paths focused solely on diagnostic accuracy are insufficient
to meet clinical needs, emphasizing the necessity to establish a
``safety-first'' evaluation and optimization
paradigm\cite{ref16}.

Notably, the domain-specific medical model MedGPT maintains a balanced,
high-level performance across both dimensions, whereas all generalist
LLMs score lower on safety relative to effectiveness. This divergence
highlights the urgent need for targeted optimization of general LLMs
through algorithmic enhancements that prioritize safety thresholds,
augmented training datasets incorporating high-risk clinical decision
trees, and integration of risk alerting mechanisms to ensure reliability
in patient-facing applications\cite{ref16}. Furthermore, all
models show a marked 13.3\% performance decline in high-risk scenarios
compared to ordinary cases (p $<${} 0.0001), aligning with
previous findings that LLMs perform worse in dynamic open-ended clinical
dialogues than in static testing environments\cite{ref20}. These
findings expose systemic shortcomings in current LLMs' clinical
knowledge depth, emergency reasoning, and risk alert systems when
confronting life-threatening situations, underscoring the necessity of
CSEDB's ``risk-weighted stratification'' design. By quantifying the
impact of high-risk tasks on overall scores, this approach compels
models to prioritize the enhancement of critical safety
competencies\cite{ref30}.

In summary, the innovations of the CSEDB include: first, the
construction of 30 indicators based on clinical expert consensus that
capture real-world risk consequence weighting; second, the use of
open-ended question-answering formats to simulate authentic clinical
interactions, thereby overcoming the scenario distortion limitations of
closed tasks like multiple-choice MedQA; third, a combination of
automated scoring and manual verification to balance evaluation
efficiency with accuracy. Together, these design choices establish a
clinically interpretable and standardized benchmark that maps technical
metrics onto clinical utility, providing actionable tools for
cross-model comparison and regulatory assessment of medical LLMs.

Regarding model improvement directions, identified weaknesses in
specific indicators---such as low scores on the scientific validity of
combination therapy plans (E13 $\leq$ 0.6) and rationality of follow-up plans
(E09 $\leq$ 0.8)---highlight priorities for enhancement. Developers should
focus on strengthening drug safety databases, optimizing decision logic
for patients with multiple comorbidities, and training with simulated
high-risk scenarios to boost emergency decision-making
capabilities\cite{ref31}. The practical value of prompt
engineering is also demonstrated: structured prompts significantly
improve safety and effectiveness scores (p $<${} 0.01), offering a
cost-effective pathway to optimize existing models by standardizing
output frameworks, such as enforcing risk alert modules, to rapidly
mitigate clinical application risks\cite{ref31}.

Nevertheless, this study has limitations. First, despite covering 26
specialty departments, the inclusion of rare diseases and multimodal
inputs such as imaging and laboratory results remains insufficient,
potentially limiting comprehensiveness\cite{ref32}. Second, data
diversity is constrained by reliance on single-turn text-based
interaction, which does not replicate the multi-turn nature of real
patient-provider communication, potentially underestimating real-world
model biases\cite{ref20}

. Future work should incorporate richer evaluation dimensions. Finally,
the assessment focuses solely on Chinese clinical question-answering
scenarios without cross-linguistic validation across multiple countries.
Expanding to multilingual clinical contexts will not only broaden
linguistic coverage but also explore variations in medical concepts and
communication patterns, thereby enhancing model generalizability and
adaptability.

In conclusion, this study introduces the CSEDB as an innovative
framework revealing critical shortcomings of current LLMs in healthcare.
It underscores the imperative for clearly defined task boundaries and
validated model reliability in medical applications. By prioritizing
high-risk scenario risk control and specialty knowledge depth through
interdisciplinary collaboration among clinicians, AI researchers, and
ethicists, continuous refinement of evaluation systems will drive the
evolution of LLMs from ``assistive tools'' to ``trusted clinical
partners,'' ultimately achieving safe and effective AI-assisted clinical
care.

\section{Data Availability}

All the Supplymentary Tables and Appendix Tables used in the study are
also available in the following
repository: \url{https://github.com/Medlinker-MG/CSEDB}

\section{Code Availability}

All code for reproducing our analysis is available in the following
repository: \url{https://github.com/Medlinker-MG/CSEDB}

\section{Acknowledgements}

We would also like to thank all the physicians for their contributions.
This work would not have been possible without the insight and
generosity of the physicians who contributed their time and expertise to
CESD Benchmark.

\section{Author Contributions}

YY, NL and JW designed and supervised the study. ZT, HY, QG, YJ, LM, YT,
YG established clinical safety and effectiveness evaluation metrics, SW,
WS and ZL established clinical data standardization and risk modeling,
HM, ZH, RL, MC, YL, DY, HG and ML contributed to clinical data
collection. SW, TG, YW, KM, HM and ZH performed the experiment. HM, ZH,
LM, WS, YJ, YT, CW, YG, QY, RL, MC, LN, ZW, PY, ML, YL, HZ, HS, LC, QZ,
SL, LZ, HG, DY, LM and YY generated benchmark databases, formulating
standards, as well as reviewing and revising them. YW, TG, LZ and WS
performed data analysis, and figure preparation. HJ, ST, SZ, CZ provided
technical support. SW, TG, YW and KM drafted the manuscript. ZT, HY, QG
and YJ,contributed to the revision. KM arranged figures and drew
illustrations. All authors had full access to all the data in the study,
discussed the results, and accepted the responsibility to submit the
final manuscript for publication. All authors have read and approved the
final version of the manuscript.

\section{Conflict of Interest}

SW, TG, YW, WS, ZL, KM, DY, HG and LM are employees of Medlinker
Intelligent and Digital Technology Co., Ltd, Beijing, China. All other
authors have declared no conflicts of interest.

\section{Funding}

The authors have no funding sources to declare.

\section{Ethics Statement}

All data sources we use to construct the Clinical Safety-Effectiveness
Dual-Track Benchmark, CSEDB benchmark dataset are publicly available and
free to use without copyright infringement. All questions in the CSEDB
dataset have been appropriately anonymized so that they do not contain
sensitive private information about patients. We do not foresee any
other possible negative societal impacts of this work.

\section{Additional Information}

Extended data and appendix table is available for this paper at
\url{https://github.com/Medlinker-MG/CSEDB}

\section*{Supplementary}

\begin{figure}[H]
  \centering
  \includegraphics[width=5.76806in,height=6.99375in]{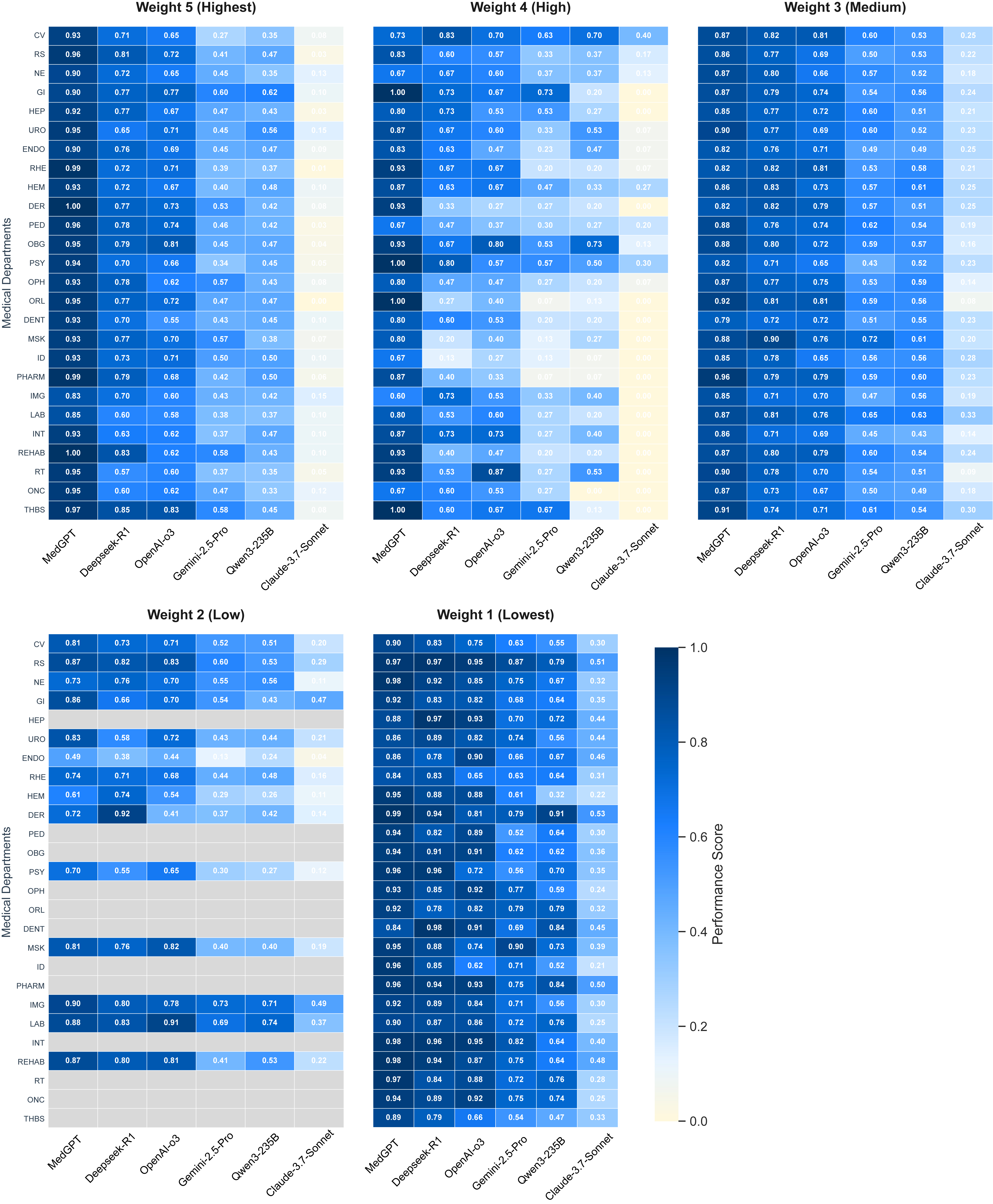}
  \captionsetup{labelformat=empty, font=small}
  \caption{\textbf{Figure S1. Comparison of LLM performance by department and weighted category.}}
  \label{fig:images1}
\end{figure}

Each row represents a department (abbreviated). The five panels correspond to weight categories from high to low. Departments without test questions in specific weight categories are marked in grey as NA. Weighted scores for each condition are labeled on the figure. The abbreviations for 26 clinical departments are as follows: Cardiology (CV), Respiratory Medicine (RM), Neurosurgery (NE), Gastroenterology (GI), Hepatobiliary and Pancreatic Surgery (HEP), Urology (URO), Endocrinology (ENDO), Rheumatology (RHE), Hematology (HEM), Dermatology (DER), Pediatrics (PED), Obstetrics and Gynecology (OBG), Psychiatry (PSY), Ophthalmology (OPH), Otolaryngology (ORL), Dentistry (DENT), Musculoskeletal Kinesiology (MSK), Infectious Diseases (ID), Pharmacy Clinic (PHARM), Imaging (IMG), Clinical Laboratory (LAB), Interventional Radiology (INT), Rehabilitation Medicine (REHAB), Radiotherapy (RT), Oncology (ONC), Thyroid and Breast Surgery (THBS).

\begin{figure}[H]
  \centering
  \includegraphics[width=5.76806in,height=3.95625in]{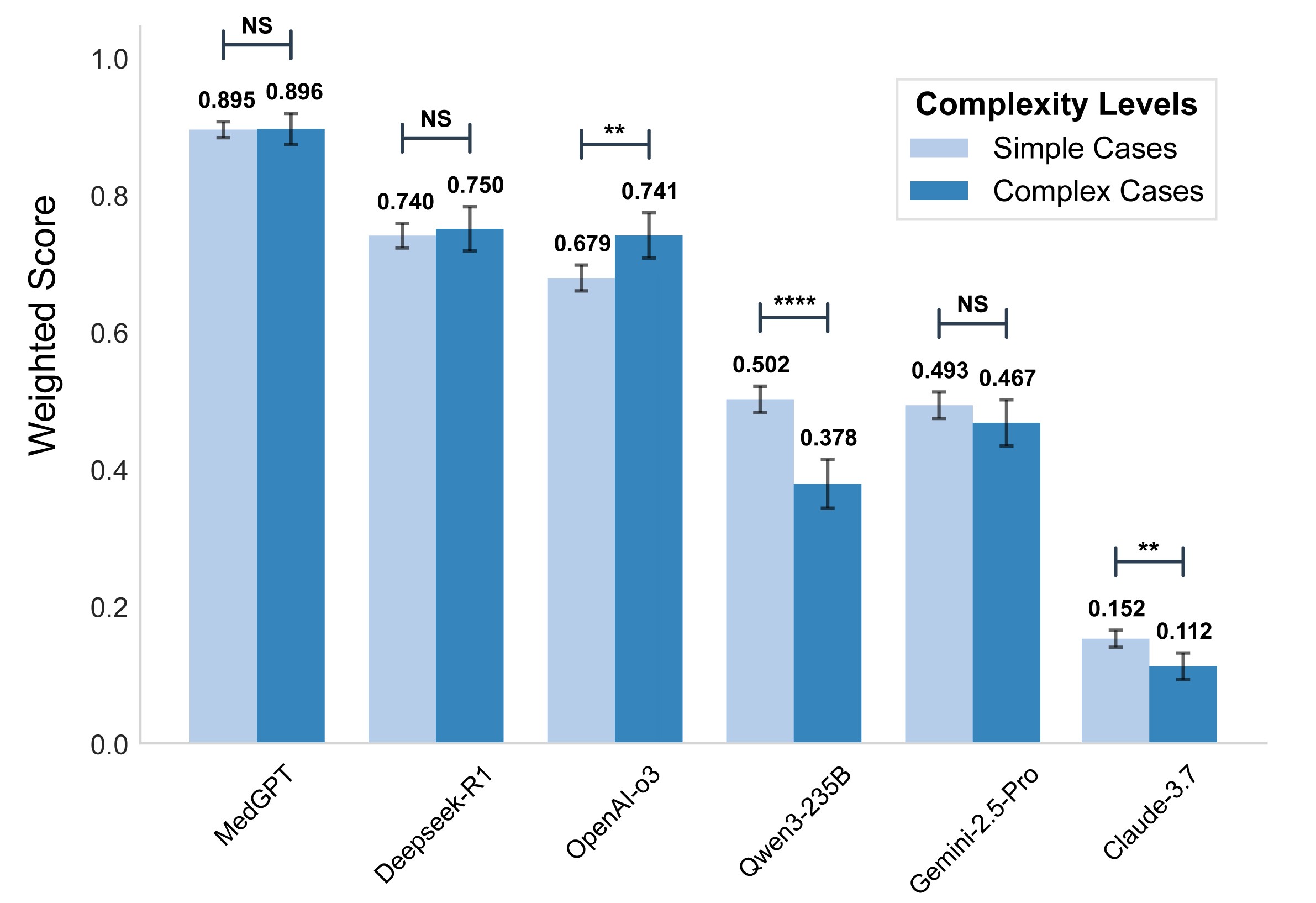}
  \captionsetup{labelformat=empty}
  \caption{\textbf{Figure S2. Performance of LLMs Across Case Complexity Levels.}
Evaluates the performance of six LLMs on~simple cases~(light blue)
and~complex cases~(dark blue) using bootstrap analysis, with error bars
representing 95\% confidence intervals. Statistical significance (NS =
non - significant; *p $<${} 0.05; **p $<${} 0.01; ****p
$<${} 0.0001)}
  \label{fig:images2}
\end{figure}

\end{document}